\documentclass[10pt,letterpaper]{article}
\usepackage[a4paper,
            bindingoffset=0.2in,
            left=1in,
            right=1in,
            top=1in,
            bottom=1in,
            footskip=.25in]{geometry}
\usepackage{tabularx,booktabs}
\usepackage{url}
\usepackage{fullpage}
\usepackage{booktabs}       
\usepackage{amsfonts}       
\usepackage{graphicx}
\usepackage{times}
\usepackage{nicefrac}       
\usepackage{microtype}      
\usepackage{lipsum}
\newcommand{\junk}[1]{}
\usepackage{graphicx}
\usepackage{float} 
\usepackage{subfigure}
\usepackage{threeparttable}
\usepackage{stfloats}
\usepackage{makecell}
\usepackage{appendix}
\usepackage{authblk}
\usepackage[colorlinks,linkcolor=blue]{hyperref}

\begin{document}
\thispagestyle{empty}

\title{HMD-AMP: Protein Language-Powered Hierarchical Multi-label Deep Forest for Annotating Antimicrobial Peptides}

\author[1,2,3]{Qinze Yu}
\author[1]{Zhihang Dong}
\author[2,3]{Xingyu Fan}
\author[1]{Licheng Zong}
\author[1,2]{Yu Li \thanks{Corresponding Author. Email: liyu@cse.cuhk.edu.hk}}
\affil[1]{\small Department of Computer Science and Engineering, CUHK, Hong Kong SAR, China}
\affil[2]{\small The CUHK Shenzhen Research Institute, Hi-Tech Park, Nanshan, Shenzhen, 518057, China}
\affil[3]{\small University of Electronic Science and Technology of China, Chengdu, Sichuan, China}

\date{}

\maketitle
\begin{abstract}
  Identifying the targets of an antimicrobial peptide is a fundamental step in studying the innate immune response and combating antibiotic resistance, and more broadly, precision medicine and public health. There have been extensive studies on the statistical and computational approaches to identify (i) whether a peptide is an antimicrobial peptide (AMP) or a non-AMP and (ii) which targets are these sequences effective to (Gram-positive, Gram-negative, etc.).  
Despite the existing deep learning methods on this problem, most of them are unable to handle the small AMP classes (anti-insect, anti-parasite, etc.). And more importantly, some AMPs can have multiple targets, which the previous methods fail to consider.  
In this study, we build a diverse and comprehensive multi-label protein sequence database by collecting and cleaning amino acids from various AMP databases.
To generate efficient representations and features for the small classes dataset, we take advantage of a protein language model trained on 250 million protein sequences. 
Based on that, we develop an end-to-end hierarchical multi-label deep forest framework, HMD-AMP, to annotate AMP comprehensively. After identifying an AMP, it will further predict what targets the AMP can effectively kill from eleven available classes. 
Extensive experiments suggest that our framework outperforms state-of-the-art models in both the binary classification task and the multi-label classification task, especially on the minor classes. Compared with the previous deep learning methods, our method improves the performance on macro-AUROC by 11\%. The model is robust against reduced features and small perturbations and produces promising results. 
We believe HMD-AMP will both contribute to the future wet-lab investigations of the innate structural properties of different antimicrobial peptides and build promising empirical underpinnings for precise medicine with antibiotics. 

\textbf{Keywords}: antimicrobial peptides, deep forest, protein language model, multi-label classification 
\end{abstract}

\newpage
\section{Introduction}
Antimicrobial peptides (AMPs) are potent, broad-spectrum antibiotics, which can help us combat diseases such as bacterial infections.
For example, they have been found and synthesized to combat \emph{pseudomonas aeruginosa} \cite{mwangi2019antimicrobial}, to heal wounds \cite{thapa2020topical}, and to even potentially work against coronavirus \cite{elnagdy2020potential}.   Meanwhile, natural antimicrobial peptides have become an exciting area of research over the recent decades due to rather an inevitable risk associated with antibiotics:  
some antimicrobials can bring detrimental effects on the body’s normal microbial content by indiscriminately attacking both the pathological occurring and beneficial ones, which 
damage the essential functions of our lungs, intestines and other organs. These side-effects root from the broad-spectrum property of some antibiotics, which could cause serious repercussions on immunity, nutrition and worse still, leading to a relative overgrowth of certain bacteria and fungi \cite{rafii2008effects,price2012staphylococcus,solomon2014antibiotic,world2014antimicrobial}. The latter repercussion could further lead to secondary infection such as \textit{clostridioides difficile}, which stems from the overgrowth of microorganisms that are antibiotic-resistant \cite{saha2019increasing}. 
Consequently, to alleviate the harm caused by antibiotics, inferences on the target of AMPs and how different peptide sequences kill targets such as viral pathogens and gram-positive bacteria are vital for major progress in antimicrobial peptide research and full utilization of AMP functions.

Scientific contributions to antimicrobial peptide research include a wide range of wet-lab studies and computational biology studies. Examples of the former include finding out novel AMPs such as SAAP-148 that combats drug-resistant bacteria and biofilm \cite{de2018antimicrobial} and LL-37 that works against \emph{staphylococcus aureus biofilm} \cite{kang2019antimicrobial}, extracting antimicrobial from tropical fruits \cite{thapa2020topical} and studying lipid and metal nanoparticles for antimicrobial peptide delivery \cite{makowski2019advances}.  While wet-lab research is crucial for knowledge discoveries in this domain, their limited generality considering the required time investment for each analysis makes it difficult to evaluate AMPs at scale. With the advance of statistical and computational methodologies, we have observed exciting recent progress on this challenge. These contributions can be divided into three categories. First and foremost, there are improving data availability, such as DBAASP \cite{pirtskhalava2016dbaasp}, LAMP \cite{zhao2013lamp}, CAMP \cite{thomas2010camp}, APD-3 \cite{wang2016apd3}, DAMPD \cite{seshadri2012dampd} and DRAMP 2.0 \cite{kang2019antimicrobial}. The abundance of data and computational resources enables the scientific community to train large-scale models. Second, there are computational design and syntheses efforts using methods like semi-supervised learning \cite{das2018pepcvae}. Finally, we have observed an accelerated growth of efforts on computational and statistical approaches analyzing AMPs, including statistical inferences like AntiBP2 \cite{lata2010antibp2}, propensity score-based binary response models \cite{randou2013binary} and novel multi-level pseudo-amino acid composition in iAMP-2L \cite{xiao2013iamp} using fuzzy k-nearest neighbors. In recent years, AMP classification becomes a question of interest in the machine learning community, too. The discussion over the possibility of identifying novel antibacterial peptides using chemoinformatics and machine learning can be dated as early as 2009 \cite{fjell2009identification}. As the machine learning toolkit expands and computational resources become more affordable, the methods applied in this question also become more diverse. For example, one study uses random forest in their AmPEP framework to predict antimicrobial peptides using distribution patterns of amino acids \cite{bhadra2018ampep}; other studies use a deep regression model to perform antimicrobial peptide design \cite{witten2019deep}. There are even studies using deep generative networks and molecular dynamics simulations to speed up antimicrobial peptide discoveries \cite{das2021accelerated}. Perhaps one of the most notable development that sparked our interests leveraged the power of convolutional neural networks (CNN) and long-short-term-memory networks (LSTM) \cite{veltri2018deep} for the task of antimicrobial peptides classification. The paper provides a novel use of deep learning methodologies on large-scale AMP databases to identify the target of antimicrobial peptides.

At the same time, there are several major challenges with computational approaches. Most antimicrobial databases include only sequences that are antimicrobials (positive labels), meaning that generating negative samples is challenging. Although there were efforts synthesizing some negative samples in previous studies, the “negative” cases are constructed in a way that is too easy to classify, which does not reflect the true intrinsic structures of non-AMPs. In addition, most antimicrobial databases include only gram-positive and gram-negative cases. In situations where multi-label data are provided, the class distribution tends to go extremely imbalanced. With respect to available models, traditional methodologies tend to overfit and fail to deliver similar performance on other homogeneous tasks. As a result, model performances of existing frameworks tend to decrease noticeably as the complexity of class distribution increases.

In this study, we address these problems by proposing a novel, end-to-end deep learning framework, HMD-AMP. We curated a challenging dataset that more closely aligns with the structural diversity of AMPs and non-AMPs. Our architecture comprises the following major components: an embedding layer of protein sequences, a protein language encoder, a feature transformer and a hierarchical deep forest framework making binary classifications (AMP or non-AMP) and multi-label classifications. Our framework is then compared with other state-of-the-art models in a binary task (Task 1: classification of AMP/non-AMP) and a multi-label task (Task 2: classification of effectiveness among 11 possible antimicrobial targets). Our framework outperforms all SOTA models in both tasks. We then evaluate the performance with an ablation study and a reduced feature test, and our findings are robust against data and feature perturbations.

The rest of this paper is organized as follows. We start by a more thorough overview of our end-to-end framework  and an evaluation of the associated ‘model problem’ and the ‘data problem’ with previous approaches in greater details in Section 2. We then compare our method side-by-side with many state-of-the-art approaches in section 3, followed by an introduction of our dataset and our experiment results, which include an ablation study, and a sensitivity analysis with reduced model sizes for evaluating the model robustness. Finally, we elaborate the observed limitations and potential future work based on our findings in Section 4 with a short discussion of potential applications.

\section{Methods}
  \subsection{Overview of HMD-AMP}
\begin{figure}[!hb] 
\centering 
\includegraphics[width=0.8\textwidth]{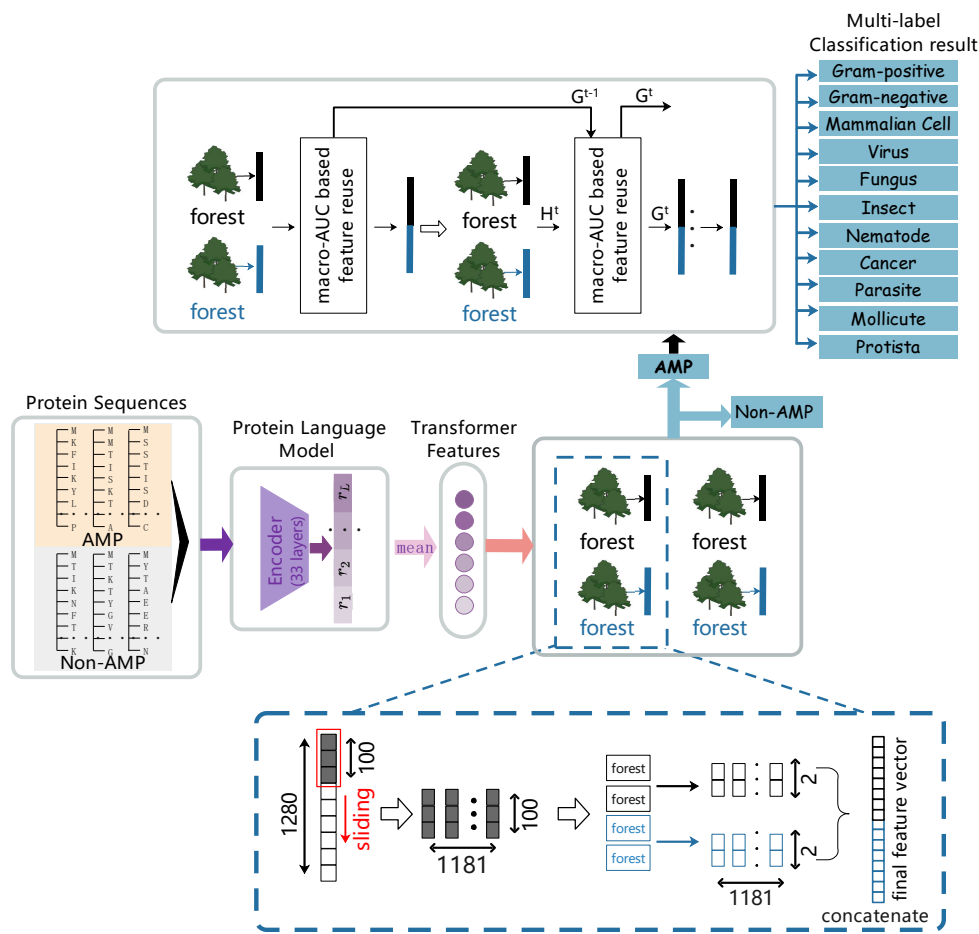} 
\caption{Overview of HMD-AMP. Top panel: HMD-AMP is consisting of one feature extraction model as well as two prediction models. The feature extraction model processes protein sequences into feature vectors. The first prediction model's inputs are feature vectors, and the model predicts a protein is an AMP or not. If the protein is an AMP, the second prediction model predicts its 11 biological functions (multi-label classification). Bottom panel: the way sliding windows scan the features.} 
\label{F1} 
\end{figure}

HMD-AMP is a supervised machine learning framework consisting of one feature extraction model as well as two prediction models. Given a protein sequence as input, HMD-AMP first extracts its features and then uses the generated features as the inputs to the prediction models. Our prediction models are designed with a two-level prediction strategy including a prediction model that predicts whether a given protein sequence is an AMP, and a second prediction model annotating the AMP's antimicrobial activities \cite{zou2019mldeepre,li2021hmd}. Specifically, HMD-AMP performs feature extraction and function prediction respectively, and it deploys a hierarchical structure of our AMP dataset labeling space. Accordingly, given any sequence analyzed by the HMD-AMP framework, the first model extracts the structural feature. The extracted features are then used as inputs to the prediction model to predict whether the sequence is an AMP or not. Furthermore, if the sequence is predicted as an AMP, the next prediction model predicts the sequence into 11 biological functions (see Figure \ref{F1}), and biological functions of an AMP are virtually its target groups. This hierarchical framework aims to use the detailed structural information of AMPs to improve the accuracy of prediction model results and to help alleviating the data imbalance problem.

For the feature extraction part of HMD-AMP, the model is an ESM-1b Transformer \cite{rives2019biological}, which takes the raw sequences as inputs. Then, the outputs of the feature extraction model are used as the inputs for the following deep forest \cite{zhou2017deep} model. 
The structure of the hierarchical classification model is illustrated in Figure \ref{F1} (top panel). 
Section \ref{sec22} elaborates each component of our proposed model structure.

\subsection{Deep learning model}
\label{sec22}
At the feature extraction level, the model is an ESM-1b transformer \cite{rives2019biological}: a transformer-based self-supervised protein language model trained on the UniRef UR50/50 database \cite{suzek2007uniref}. ESM-1b processes inputs as character
amino acids sequences, using positional embeddings instead of making assumptions on the ordering of the input.
From ESM-1b, we obtain residue-level sequence embeddings. In order to get the protein-level embeddings as the inputs to the function prediction model, we average across all residue positions of residue-level sequence embeddings, hence getting a 1280-dimension feature vector for each sequence.

Transformer's \cite{vaswani2017attention} self-attention mechanism and its ability to model long-range dependencies, which reflect structural properties of protein sequences, enable themselves to predict amino acid residual contact because the attention maps generated within the Transformer naturally correspond to the information between the various residues in the sequence. 

Then, at each function prediction level, the model is a deep forest \cite{zhou2017deep} model, which demonstrates a cascade forest structure. Each level of the cascade receives the feature information processes by its previous level and feeds its outputs to the next level as inputs. Each cascade level is an ensemble of decision tree forests, and different types of forests are included to make the model diverse. In our model design, at each level, we deploy two completely-random tree forests and two random forests: for these forests, the inputs are embeddings obtained from the feature extraction model. Each random forest contains 1000 trees, it randomly selects $\sqrt{d}$ features as candidate features (\emph{d} being the number of input features) and the one with the best \emph{gini} value is chosen for the split. For 
our 11 labels in the multi-label classification, we compute their \emph{gini} values as follows:
\begin{equation}
\mathbf{Gini}(p)= \sum_{k=1}^{11}2p_{k}(1-p_{k}), \label{eq1}
\end{equation}
where $p_{k}$ is the probability that the sample has the label \emph{k}.
Every complete-random tree forest also contains 1000 completely random trees, which are designed to detect important motifs across the inputs. 

To enhance the model's ability to handle feature relationships, a multi-grained scanning procedure is designed as shown in Figure \ref{F1} (bottom panel). Specifically, sliding windows scan the given input features. Our inputs are 1280-dimension raw feature vectors, and a window size of 100 is used. For sequence data, a 100-dimension feature vector will be generated by sliding the window for one feature. As a result, a total of 1181 feature vectors are produced for each iteration. Feature vectors extracted from positive/negative training examples are regarded as corresponding instances, and these instances are used to train the forest and then to generate the estimated class distributions, which would be converted to class vectors. Finally, the class vectors are concatenated as transformed features. Take the binary classification task as an example: we have 2 classes, and 1181 2-dimension class vectors are produced by each forest. As a result, the 9448-dimension transformed feature vector is taken as the counterpart of the original 1280-dimension raw feature vector.

In general, each level of the cascade receives feature information processed by its preceding level
and feeds its processed results to the next level as inputs until there is no significant performance gain, when the training process terminates. This process makes the deep forest appropriately determines the complexity of its model by termination. Also, deep forest does not rely on backpropagation, so it is suitable for training data with either imbalance labels or small sample sizes, hence preventing the model from overfitting. Further, two mechanisms are added to help the deep forest performs well in predicting specific antimicrobial activities. The first mechanism is a measure-aware feature reuse \cite{yang2019multi}. That is, if the confidence of the current layer is lower than the threshold determined during training, the better representation of the previous layer is partially reused. The confidence to each label is an estimation
of the respective label distributions. To accommodate labels with smaller representation, we choose macro-AUC as the confidence-computing metric (Figure \ref{F1} top panel). Macro-AUC is a label-based measure \cite{wu2017unified} that is defined as follows:
\begin{equation}
\mathbf{AUC}_{macro}(F)=\frac{1}{l} \sum_{j=1}^{l} \frac{\left|\mathcal{S}_{macro }^{j}\right|}{\left|Y_{\cdot j}^{+} \| Y_{\cdot j}^{-}\right|},
\end{equation}
which is per-class raw average of AUC, where $Y$ is the true label; $Y_{\cdot j}$ is the $j$-th column of the label matrix, and ‘+’ (‘-') is the relevant
(irrelevant) note. $\mathcal{S}_{{macro}}$
is the set of correctly ordered instance
pairs on each label:
\begin{equation}
\mathcal{S}_{macro }^{j}=\left\{(a, b) \in Y_{\cdot j}^{+} \times Y_{\cdot j}^{-} \mid f_{j}\left({x}_{a}\right) \geq f_{j}\left({x}_{b}\right)\right\},
\end{equation}
where the $f_{ij}$ means the confidence score of $i$-th instance on $j$-th label,

Therefore, the confidence computing method of each label is shown as Equation \ref{eq2}. Here, \emph{m} refers to the number of sequences and \emph{$p_{ij}$} means Pr[$\hat{y}_{ij}$=1].
\begin{equation}
\mathbf{confidence}_{(j)}=\sum_{i=0}^{m}\prod_{k=1}^{i}p_{kj}\prod_{k=i+1}^{m} (1-p_{kj}). \label{eq2}
\end{equation}

The second mechanism is the measure-aware layer growth, which focuses on the learning of representation, and it efficiently enhances the representation through various measures while reducing overfitting and controlling model complexity. For the cascade forest, we artificially set the maximal depth of the layers as 20 in the initialization step. If the model has grown to the maximal number, the training process terminates. In the initialization step, we also initialize the performance vector, which records the performance value on training data in each layer. Here, we still choose macro-AUC as the measure to indicate performance. During each layer $t$, the forest is fitted according to the training data so that we get this layer's classifier $h_{t}$. With the classifier, we predict the representation $H_{t}$ (Equation \ref{eq3}), where $\mathrm{X}$ is the training data, and $\mathrm{G}_{t-1}$ is the representation of the last layer. Then we obtain the new representation of the current layer by measure-aware feature reuse. Due to the layer growth being measure-aware, the model needs to compute the macro-AUC after fitting each layer. When the measure is not getting better in recent three layers, an early stopping mechanism forces the deep forest to stop growing even if it yet reaches the maximum depth. At the same time, the best performance layer index is recorded. Therefore, with such well-designed mechanisms, multi-label deep forest \cite{yang2019multi} is very appropriate for solving multi-label problems.
\begin{equation}
\mathrm{H}_{t}=h_{t}\left(\left[\mathrm{X}, \mathrm{G}_{t-1}\right]\right).
\label{eq3}
\end{equation}

\section{Results}
  In this section, we analyze our model in comparison with several published, state-of-the-art methods and train them with our dataset. In particular, we start by describing the different settings of the experiments and evaluating our proposed model compared to these SOTA models.
\subsection{Alternative methods}
We first introduce the benchmark result produced by Veltri et al. \cite{veltri2018deep} combining CNN and LSTM for antimicrobial recognition. In addition, the development of meta learning makes
Model-Agnostic Meta-Learning (MAML) \cite{finn2017model} a suitable candidate. Instead of learning a model that can be used directly for prediction, such meta-learning methods learn how to learn a model faster and better instead. Meanwhile, we include the \emph{Probabilistic Model-Agnostic Meta-Learning} \cite{finn2018probabilistic}, an extension of the original MAML, as another candidate. We encourage interested readers to refer to their initial manuscripts for details beyond our brief summary. MAML can be interpreted as approximate inference for the posterior \cite{grant2018recasting}. It uses the \emph{maximum a posteriori} (MAP) value. The algorithm evaluates the variational lower-bound for the logarithm of the approximate likelihood, which can be written as 
\begin{equation}
\log p\left(\mathbf{y}_{i}^{{test}} \mid \mathbf{x}_{i}^{{test}}, \mathbf{x}_{i}^{\mathrm{tr}}, \mathbf{y}_{i}^{ {tr }}\right) \geq E_{\theta \sim q_{\psi}}\left[\log p\left(\mathbf{y}_{i}^{ {test }} \mid \mathbf{x}_{i}^{{test }}, \phi_{i}^{\star}\right)+\log p(\theta)\right]+\mathcal{H}\left(q_{\psi}\left(\theta \mid \mathbf{x}_{i}^{{test }}, \mathbf{y}_{i}^{{test }}\right)\right).
\end{equation}
In this bound, it essentially performs approximate inference via MAP on $\phi_{i}$ to obtain $p\left(\phi_{i} \mid \mathbf{x}_{i}^{\mathrm{tr}}, \mathbf{y}_{i}^{\mathrm{tr}}, \theta\right)$, and uses the variational distribution for $\theta$ only. Then, the inference network is given by 

\begin{equation}
q_{\psi}\left(\theta \mid \mathbf{x}_{i}^{{test}}, \mathbf{y}_{i}^{{test}}\right)=\mathcal{N}\left({\mu}_{\theta}+\gamma_{q} \nabla \log p\left(\mathbf{y}_{i}^{{test}} \mid \mathbf{x}_{i}^{{test}},{\mu}_{\theta}\right) ; \mathbf{v}_{q}\right).
\end{equation}
The training is performed by backpropagating gradients, and this process includes a term for the likelihood 
$\log p\left(\mathbf{y}_{i}^{ {test }} \mid \mathbf{x}_{i}^{ {test }}, \mathbf{x}^{\mathrm{tr}}, \mathbf{y}^{ {tr }}, \phi_{i}^{\star}\right)$ and the KL-divergence between the sample \emph{$\theta$} $\sim$ $q_{\psi}$ and the prior $p(\theta)$.
We try to implement both MAML and PMAML against CNN-LSTM \cite{veltri2018deep} benchmark. Indeed, the parameters randomly initialized by PMAML are hard to train, and such PMAML model only performs well if we use parameters obtained from the trained MAML model for the training of PMAML.

Another approach to highlight is the AMAP \cite{gull2019amap}, which is a hierarchical multi-label prediction model that annotates the biological functions of AMP sequences, using extreme Gradient Boosting (XGBoost) \cite{chen2016xgboost}. XGBoost is based on boosted trees, and it learns by minimizing the objective function:
\begin{equation}
L(\Phi)=\sum_{i} l\left(\hat{\mathrm{y}}_{\mathrm{i}}, \mathrm{y}_{i}\right)+\sum_{k} \Omega\left(f_{k}\right),
\end{equation}
where 
$\Omega\left(f_{k}\right)=\gamma T+\frac{\lambda}{2}\left\|u_{k}\right\|$. Here, $l\left(\hat{\mathrm{y}}_{\mathrm{i}}, \mathrm{y}_{i}\right)$ is the loss function of predicted model output $\hat{\mathrm{y}}_{\mathrm{i}}$
and actual output $\mathrm{y}_{i}$ for all examples. $\Omega\left(f_{k}\right)$ is a regularization
function that is based on the number of trees $T$ and the norm of the
vector of scores $u$ at the $k$-leaf of the trees. The regularization
parameters $\gamma$ and $\lambda$ control the relative contribution of the two regularization
factors in contrast to the minimization of the loss function. We include AMAP in our comparisons in Section \ref{sec34}.

\subsection{Datasets}
We compile a comprehensive multi-label AMP database with a high degree of confidence. 
Specifically, we collect and clean amino acids sequences from three published AMP databases: Database of Antimicrobial Activity and Structure of Peptides (DBAASP) \cite{pirtskhalava2016dbaasp}, an update to LAMP database linking antimicrobial peptide (LAMP2) \cite{ye2020lamp2}, and data repository of antimicrobial peptides (DRAMP) \cite{shi2021dramp}. Then, we remove the identical and duplicate sequences from our database. The resulting database is composed of 18514 high-quality sequences, coupled with labels of 11 antimicrobial activities classes (Table \ref{table 1} and Table \ref{table 8}). In the design of our machine
learning models, these sequences are taken as positive examples.

\begin{table*}
\begin{minipage}[t]{0.48\linewidth}
\begin{center}
\centering
\caption{Target groups and the number of peptides in each
category in our positive dataset.}
\label{table 1}
\begin{tabular}{ll}
\toprule
Activity & Peptides Count\\
\midrule
Gram-positive & 11486\\
Gram-negative & 11958\\
Mammalian Cell & 7403\\
Virus & 3779\\
Fungus & 5514\\
Insect & 181\\
Cancer & 2271\\
Parasite & 405\\
Mollicute & 31\\
Nematode & 34\\
Protista & 41\\
\bottomrule
\end{tabular}
\end{center}
\end{minipage}
\begin{minipage}[t]{0.48\linewidth}
\begin{center}
\begin{threeparttable}
\centering
\caption{Peptides' label amount in our positive dataset.}
\label{table 8}
\begin{tabular}{ll}
\toprule
No of labels & Peptides Count\\
\midrule
1 & 6149\\
2\tnote{a} & 4246\\
3 & 4861\\
4 & 2480\\
5 & 711\\
6 & 65\\
7 & 2\\

\bottomrule
\end{tabular}
\begin{tablenotes}
        \footnotesize
        \item[a] Here 2 means a peptide has 2 labels (biological functions)  
      \end{tablenotes}
\end{threeparttable}
\end{center}
\end{minipage}
\end{table*}

We extract 12659 peptides with the highest BLAST \cite{madden2013blast} similarity scores against the AMPs in our multi-label AMP database from Uniprot \cite{uniprot2019uniprot} and these peptides show no antimicrobial activity. Specifically, to avoid sequence and composition biases that affect our machine learning, we filter peptides using an approach similar to previous work \cite{veltri2018deep} and remove all peptides with more than 40$\%$ sequence identity to each other  
using CDHIT \cite{fu2012cd}. It leaves a total of 8534 peptides and we use these as negative peptides to train and evaluate our model. We use 5-fold stratified cross-validation to evaluate the performance of our model, the dataset would randomly be divided into five folds, and at each time, four of them are chosen for the model training and the remaining one fold is used to test the trained model. As a result, average results are generated from repeating the above procedure five times. 

\subsection{Implementation details}
We use the Tensorflow toolkit to write our code and train HMD-AMP with 2 NVIDIA GeForce RTX 3090 GPUs. We train the first deep forest model on the whole dataset. Because our negative set bears a strong resemblance to the positive set, deep forest is forced to learn to be a more powerful model. Then, we perform the next level multi-label deep forest model with our positive set, the dataset contains more than 18000 AMP sequences, and each of them has 11 labels, indicating which target groups the sequence resists.
When training the models, we first input the protein sequence directly into ESM-1b model, and for each sequence, we can obtain a 1280-dimension embedding vector, which is the result of averaging across all residue positions of residue-level sequence embeddings. Such an embedding vector is then fed into a deep forest model, with two random forests and two complete-random tree forests with 1000 completely random trees for the binary classification. If one sequence is predicted to be an AMP, the multi-label deep forest annotates its target groups, this model has the same tree structure with the binary classification model.

\begin{table}[t]
\centering
\caption{The AMP/non-AMP classification results between
different methods}
\label{table 2}
\begin{threeparttable}

\begin{tabular}{lllll}
\hline
~                       & Accuracy       & Precision\tnote{a}      & Recall         & F1-score       \\ \hline
AMAP                    & 0.882          & 0.882          & 0.851          & 0.863          \\
DL                      & 0.956          & 0.951          & 0.944          & 0.947          \\
MAML+DL                 & 0.941          & 0.935          & 0.939          & 0.937          \\
PMAML+DL                & \textbf{0.961} & \textbf{0.957} & 0.951          & 0.954          \\
HMD-AMP                  & 0.956          & 0.955          & \textbf{0.953} & \textbf{0.954} \\ \hline
\end{tabular}
 \begin{tablenotes}
        \footnotesize
        \item[a] The precision, recall, and F1-score are the macro averages over the AMP and Non-AMP  
      \end{tablenotes}
\end{threeparttable}
\end{table}

\subsection{Performance comparison}
\label{sec34}
Here, we evaluate the models' performance we proposed above. We refer to Veltri et al.'s work \cite{veltri2018deep} as DL, and DL model trained by MAML \cite{finn2017model} and PMAML \cite{finn2018probabilistic} algorithms are called MAML+DL and PMAML+DL respectively.  
We use a 5-fold stratified cross-validation to evaluate
the performance of HMD-AMP. In this experiment,
we randomly divide our dataset into five folds.
Each time, we choose four folds from the dataset for the
model training and test the trained model on the
remaining one. To avoid data bias, average results are
generated from repeating the above procedure five
times. 
\subsubsection{Binary classification performance comparison}
As shown in Table \ref{table 2}, for the AMP/non-AMP classification, HMD-AMP shows higher recall (0.953) and F1-score (0.954) with promising accuracy (0.956) and precision (0.955) than existing methods.
One observation is that deep network methods generally have better performances on binary (AMP/non-AMP) classification than AMAP. One of the reasons is when there is enough data for training, deep models can often fit better functions for prediction, whereas AMAP is considered a traditional machine learning model.

\begin{table}[b]
\centering
\caption{The AMP biological functions (multi-label) classification results
between different methods}
\label{table 3}
\begin{threeparttable}

    \begin{tabular}{lllll}
    \hline
    ~        & Accuracy & Precision\tnote{a} & Recall & macro-AUC \\ \hline
    AMAP     & 0.913    & 0.704     & 0.695  & 0.859         \\
    DL       & 0.904    & 0.536     & 0.451  & 0.844      \\
    MAML+DL  & 0.909    & 0.794     & 0.671  & 0.861      \\
    PMAML+DL & 0.927    & 0.846    & 0.748  & 0.897      \\
    HMD-AMP   & \textbf{0.958}   & \textbf{0.887}     & \textbf{0.812}  & \textbf{0.915}      \\ \hline
    \end{tabular}
 \begin{tablenotes}
        \footnotesize
        \item[a] The precision and recall are the macro averages over the 11 biological functions classes  
      \end{tablenotes}
\end{threeparttable}
\end{table}

\subsubsection{Multi-label classification performance comparison}
\begin{figure}[t] 
\centering 
\includegraphics[width=1.0\textwidth]{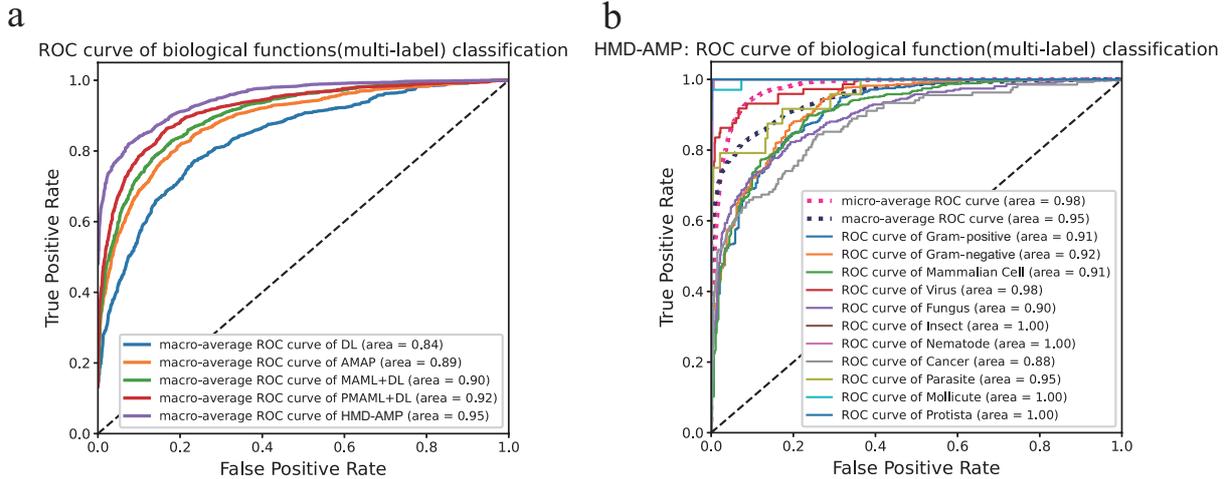} 
\caption{\textbf{a}: macro-average ROC curves comparison of 5 models. \textbf{b}: ROC curves of our model, including macro/micro-average and 11 biological function labels.} 
\label{F6} 
\end{figure}

\begin{figure}[t] 
\centering 
\includegraphics[width=0.6\textwidth]{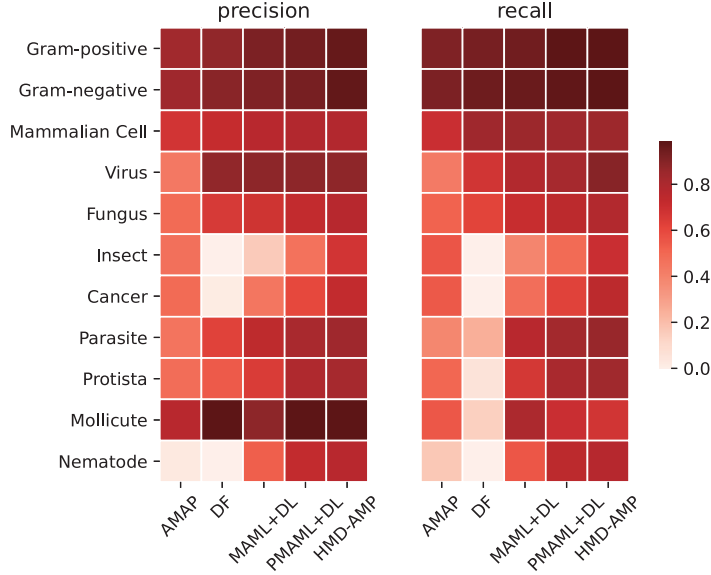} 
\caption{Detailed prediction performance comparison on each biological function label.} 
\label{F3} 
\end{figure}

HMD-AMP beats the state-of-the-art results (as shown in Table \ref{table 3} and Figure \ref{F6}) in this more challenging task.  HMD-AMP significantly outperforms across all measures including accuracy, precision, recall, and macro-AUC. 
Despite all methods having relatively high accuracy (still far behind the HMD-AMP) on the AMP biological functions classification, the simple DL method cannot correctly predict labels with only a small amount of data (Figure \ref{F3}), and our HMD-AMP outperforms DL by more than 30\% on both precision and recall scores. For example, in our dataset, only 41 peptides have resistance to Protista, DL has little effect on the classification of this label, leading to many false negatives. Actually, both DL and MAML+DL have a very inconsistent performance across different classes, especially in terms of recall. In contrast, HMD-AMP is quite stable across different classes regardless of precision or recall (Figure \ref{F3}). For AMAP, its two types of sequence-based feature representation help the method performs well on multi-label classification. Nevertheless, our method outperforms AMAP by about 10\% on both precision and recall scores. Although PMAML+DL has relatively good performance on 4 evaluation metrics (still about 5\% behind our method on both precision and recall), it is largely based on parameters derived from the trained MAML+DL. If randomly initialized parameters are used, it would be difficult for PMAML+DL to obtain good performance.

\subsection{Ablation study}
With the aforementioned experiments validating the strengths of our model in multi-label classification, we develop a new experiment to see if our method works with very little data. We select 50, 100, 200 data points from our positive dataset to train HMD-AMP, MAML+DL, and PMAML+DL, and see whether our model still delivers consistent performance. We make sure that the selected data contain positive and negative data for all 11 labels. Again, every model is validated by using a 5-fold cross-validation test reporting the performance averaged over five trials, where each trial leaves out a different 20$\%$ of the selected data as a test set to validate the performance of the model trained on the other 80$\%$ of the selected data. It is worth mentioning that our method works well even with little data: we show the result in Table \ref{table 4}. When the number of data points is 200, our method outperforms the other two models on accuracy (0.928), precision (0.841), and macro-AUC (0.916). It is not surprising that PMAML has a slightly
better recall score (0.821), because PMAML was
designed to solve few-shot problems and it can be trained to converge at great speed. When the number of data points is 100, our method outperforms the other two models on accuracy (0.925) and macro-AUC (0.860), and PMAML gets the best score on precision (0.810) and recall (0.780), and when the number of data points is 50, PMAML shows slightly higher performance on all 4 measures. Although our approach is slightly behind PMAML, it still shows favorable performance and is superior to MAML across all metrics.
The result indicates that our method is also suitable for the small sample problem.

\begin{table}[b]
\centering
\caption{sensitivity analysis results}
\label{table 4}
\begin{threeparttable}
\begin{tabular}{llllll}
\hline
                                         &          & Accuracy\tnote{c}       & Precision\tnote{b}     & Recall         & macro-AUC      \\ \hline
{50\tnote{a}}                      & MAML+DL  & 0.836          & 0.725          & 0.686          & 0.759          \\
                                         & PMAML+DL & \textbf{0.861} & \textbf{0.793} & \textbf{0.782} & \textbf{0.824} \\
                                         & HMD-AMP   & 0.849          & 0.767          & 0.741          & 0.811          \\ \hline
{{100}} & MAML+DL  & 0.910          & 0.744          & 0.701          & 0.807          \\
\multicolumn{1}{c}{}                     & PMAML+DL & 0.911          & \textbf{0.810} & \textbf{0.780} & 0.831          \\
\multicolumn{1}{c}{}                     & HMD-AMP   & \textbf{0.925} & 0.804          & 0.769          & \textbf{0.860} \\ \hline
{200}                     & MAML+DL  & 0.903          & 0.787          & 0.739          & 0.853          \\
                                         & PMAML+DL & 0.907          & 0.837          & \textbf{0.821} & 0.874          \\
                                         & HMD-AMP   & \textbf{0.928} & \textbf{0.841} & 0.802          & \textbf{0.916} \\ \hline
\end{tabular}
 \begin{tablenotes}
        \footnotesize
        \item[a] 50, 100, and 200 means the number of data points used for training and testing.
        \item[b] The precision and recall are the macro averages over the 11 biological functions classes.
        \item[c] Every model is validated by using a 5-fold cross-validation test, and the performance averaged over five trials.
      \end{tablenotes}
\end{threeparttable}
\end{table}

To evaluate the effectiveness of the feature extraction model and the prediction model respectively, we conduct the model replacement test. To investigate the efficiency of the feature extraction model, we apply deep forest and take the raw representation of the sequence, i.e., one-hot encoding, as inputs. In this experiment, we name it Deep Forest. Further, to analyze our prediction model's importance, we change the prediction model into the random forest \cite{breiman2001random} and retain our feature extraction model. The random forest has 1000 trees. Likewise, we name this method as Random Forest. Table \ref{table 5} and Figure \ref{F2}a show the experimental results of HMD-AMP, Random Forest, and Deep Forest, all three methods have high accuracy, but Random Forest has low precision and recall scores. Through inspection, we find that Random Forest could hardly recognize some labels with unbalanced data (such as Protista and Mollicute): its correct predictions of few true positive cases lead to a large number of false negatives. Even a small number of false positives could result in low precision. Besides, we find that Deep Forest performs well on unbalanced labels, which is due to the large size of trees in forests (All trees in the deep forest are averaged to generate an estimate of the distribution of classes) and the measure-aware layer growth mechanism. However, HMD-AMP has been consistently ranked as the best performance across all the measures, which indicates that our feature extraction model provides effective features. These features help deep forest using more suitable motifs to make the prediction. 

\begin{table*}[b]
\centering
\caption{ablation test results}
\label{table 5}
\begin{threeparttable}
\begin{tabular}{lllll}
\hline
              & Accuracy       & Precision\tnote{a}      & Recall         & macro-AUC     \\ \hline
Deep Forest   & 0.948          & 0.776          & 0.732          & 0.889          \\
Random Forest & 0.939          & 0.513          & 0.408          & 0.791          \\
HMD-AMP        & \textbf{0.958} & \textbf{0.887} & \textbf{0.812} & \textbf{0.915} \\ \hline
\end{tabular}
 \begin{tablenotes}
        \footnotesize
        \item[a] The precision and recall are the macro averages over the 11 biological functions classes.
      \end{tablenotes}
\end{threeparttable}
\end{table*}

\subsection{Reduced feature model analysis}

\begin{figure}[tb] 
\centering 
\includegraphics[width=0.8\textwidth]{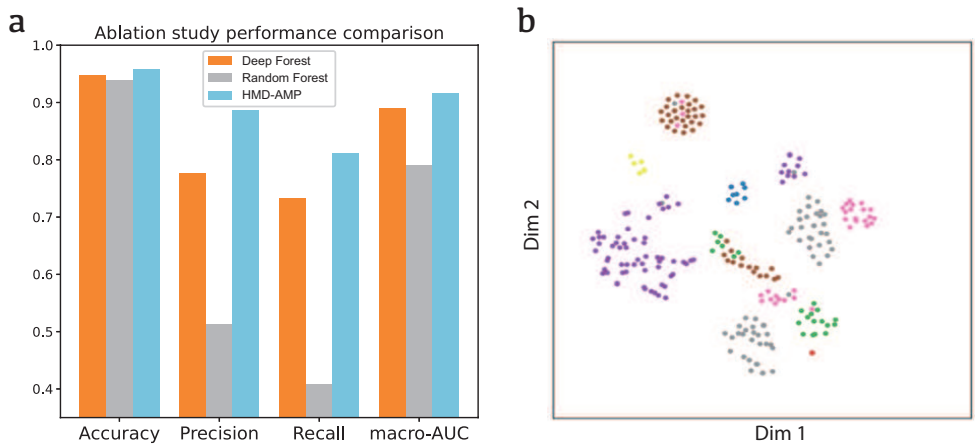} 
\caption{\textbf{a}: Ablation study of the feature extraction model and prediction model. \textbf{b}: A 2D t-SNE projection. Different colors represent AMP's different label-combination, and data points in the same cluster share a high similarity.} 
\label{F2} 
\end{figure}

To verify the similarity of input features with exactly same labels, we perform visual processing on the feature vectors. We apply t-SNE \cite{van2008visualizing}, which is a non-linear cluster recognition algorithm, for data dimensionality reduction. T-SNE finds similarity patterns in data points with multiple features, and we can see data points with high similarity converge into a cluster by projecting dimensionality reduction results onto a two-dimensional plane. We use t-SNE to reduce a 1280-dimension feature vector to a 2-dimension vector. For each label combination, we assign different numbers and use the numbers to distinguish the AMP of different labels. Some AMP sequences from our dataset are selected, and the result is shown in Figure \ref{F2}b. From the result, we clearly see differentiated clusters, which indicates the feature extraction model nicely mines the structural and functional features inside the protein sequences.

However, 1280-dimension vector features cause time-burden and memory-consumption, we try to reduce features' amount and choose valid features. we adopt Local Interpretable Model-agnostic Explanations (LIME) \cite{ribeiro2016should}. LIME is an explanation technique that explains the prediction of our classifier in an interpretable and faithful manner by learning an interpretable model locally around the prediction. LIME gets each feature's global weight (Figure \ref{F4}) by averaging the value of local prediction, and we select 48 features with the highest global weight (Figure \ref{F5}) as new features to train our prediction model. Despite using only 48 features, HMD-AMP's performance still outperforms AMAP, DL, and MAML+DL on both the AMP/non-AMP classification and the AMP biological functions classification. Reduced features (48 features) model performance has a slight drop of about 0.7$\%$ on accuracy, 0.8$\%$ on precision and recall (Table \ref{table 7}) compared with HMD-AMP using 1280 features on the AMP/non-AMP classification, but it greatly improves the training speed.

\begin{table*}[!h]
\centering 
\caption{Reduced feature model performance}
\label{table 7}
\begin{threeparttable}
\begin{tabular}{llllll}
\hline
                                             & senario of classification & accuracy\tnote{c} & precision\tnote{d} & recall & F1-score \\ \hline
{{HMD-AMP}} & binary\tnote{a}                    & 0.951    & 0.947     & 0.945  & 0.946    \\
                    & multi-label\tnote{b}               & 0.915    & 0.806     & 0.744  & 0.774    \\ \hline
\end{tabular}
\begin{tablenotes}
        \footnotesize
        \item[a] The AMP/non-AMP classification
        \item[b] The AMP biological functions classification
        \item[b] The model is validated by using a 5-fold cross-validation test, and the performance averaged over five trials
        \item[d] The precision and recall are the macro 
        averages of the AMP and Non-AMP/the 11 biological functions classes.
      \end{tablenotes}
\end{threeparttable}
\end{table*}

\subsection{Model application}
We apply our model to 20 newly synthesized antimicrobial peptides candidate sequences (Table \ref{table 6}) \cite{das2021accelerated}. In fact, only two of these sequences (YI12 and FK13) were resistant to gram-positive and gram-negative, while the other 18 were not antimicrobial peptides. We use HMD-AMP to predict 20 peptides, and find that YI12 and FK13 are predicted to have resistance to gram-positive and gram-negative. Besides, 8 of the non-AMP peptides are identified by HMD-AMP, in contrast to the method \cite{das2021accelerated} recognizing all 20 peptides as AMP. Our HMD-AMP consistently outperforms methods like AmpGram \cite{burdukiewicz2020proteomic}, AMPA \cite{torrent2012ampa}, AMPScanner \cite{veltri2018deep}, and CAMP \cite{waghu2016campr3}, which predict YI12 as non-AMP, and AMAP \cite{gull2019amap}, which predicts more non-AMP as AMP than our method.

Another interesting observation is that FK13 has the highest gram-positive (0.893, rank:1) and gram-negative (0.930, rank:1) probability among the 12 peptides that HMD-AMP predicted as AMPs, YI12 also has promising probability on gram-positive (0.799, rank:5), gram-negative (0.829, rank:4).
The result shows that our model not only predicts the existing AMP, but also carries out functional annotations for AMPs that are newly synthetic or have yet existed.

\section{Conclusion}
  We develop a hierarchical method, HMD-AMP, to facilitate the detection of antimicrobial peptides, providing detailed AMPs' biological functions annotations. Comprehensive experiments including cross-fold validation, sensitivity test, ablation study, and new peptides test, demonstrate the effectiveness and robustness of our proposed method, where our model consistently and significantly outperforms all counterparts. Most known AMP prediction methods only classify sequences into AMPs/non-AMPs, and methods that predict AMPs' biological functions are mostly based on statistics, sequence comparison, and traditional machine learning models. However, former methods' performances are worse than our model, they don't recognize labels with small amounts of data well, leading to very low recall.

Moreover, since the deep learning model trained on meta-learning algorithms performs well on little data, and it is developed to help the model learn faster in different scenarios, extending the model to do multi-tasks is a promising research direction. In the future, we will try to combine our method with the meta-learning algorithm and make it able to classify protein sequences into more groups, not limited to AMP.
Also, the 20 peptides classification result in section ‘Model application’ enlightens us to further develop HMD-AMP as a generative model, which can sample large amounts of peptides and generate new AMPs.

We believe that HMD-AMP can serve as a powerful
tool to promote the application of antimicrobial peptides and alleviate the global threat of antibiotic resistant genes. In the future,
we will incorporate other dimensions of information,
such as 3D structural information \cite{lam2019deep, wei2021protein}, into
our framework to further improve our method’s performance
and extend the application scenarios.

\bibliography{mybib.bbl}

\begin{thebibliography}{10}
\expandafter\ifx\csname url\endcsname\relax
  \def\url#1{\texttt{#1}}\fi
\expandafter\ifx\csname urlprefix\endcsname\relax\def\urlprefix{URL }\fi
\providecommand{\bibinfo}[2]{#2}
\providecommand{\eprint}[2][]{\url{#2}}

\bibitem{mwangi2019antimicrobial}
\bibinfo{author}{Mwangi, J.} \emph{et~al.}
\newblock \bibinfo{title}{The antimicrobial peptide zy4 combats
  multidrug-resistant pseudomonas aeruginosa and acinetobacter baumannii
  infection}.
\newblock \emph{\bibinfo{journal}{Proceedings of the National Academy of
  Sciences}} \textbf{\bibinfo{volume}{116}}, \bibinfo{pages}{26516--26522}
  (\bibinfo{year}{2019}).

\bibitem{thapa2020topical}
\bibinfo{author}{Thapa, R.~K.}, \bibinfo{author}{Diep, D.~B.} \&
  \bibinfo{author}{T{\o}nnesen, H.~H.}
\newblock \bibinfo{title}{Topical antimicrobial peptide formulations for wound
  healing: Current developments and future prospects}.
\newblock \emph{\bibinfo{journal}{Acta biomaterialia}}
  \textbf{\bibinfo{volume}{103}}, \bibinfo{pages}{52--67}
  (\bibinfo{year}{2020}).

\bibitem{elnagdy2020potential}
\bibinfo{author}{Elnagdy, S.} \& \bibinfo{author}{AlKhazindar, M.}
\newblock \bibinfo{title}{The potential of antimicrobial peptides as an
  antiviral therapy against covid-19}.
\newblock \emph{\bibinfo{journal}{ACS Pharmacology \& Translational Science}}
  \textbf{\bibinfo{volume}{3}}, \bibinfo{pages}{780--782}
  (\bibinfo{year}{2020}).

\bibitem{rafii2008effects}
\bibinfo{author}{Rafii, F.}, \bibinfo{author}{Sutherland, J.~B.} \&
  \bibinfo{author}{Cerniglia, C.~E.}
\newblock \bibinfo{title}{Effects of treatment with antimicrobial agents on the
  human colonic microflora}.
\newblock \emph{\bibinfo{journal}{Therapeutics and clinical risk management}}
  \textbf{\bibinfo{volume}{4}}, \bibinfo{pages}{1343} (\bibinfo{year}{2008}).

\bibitem{price2012staphylococcus}
\bibinfo{author}{Price, L.~B.} \emph{et~al.}
\newblock \bibinfo{title}{Staphylococcus aureus cc398: host adaptation and
  emergence of methicillin resistance in livestock}.
\newblock \emph{\bibinfo{journal}{MBio}} \textbf{\bibinfo{volume}{3}},
  \bibinfo{pages}{e00305--11} (\bibinfo{year}{2012}).

\bibitem{solomon2014antibiotic}
\bibinfo{author}{Solomon, S.~L.} \& \bibinfo{author}{Oliver, K.~B.}
\newblock \bibinfo{title}{Antibiotic resistance threats in the united states:
  stepping back from the brink}.
\newblock \emph{\bibinfo{journal}{American family physician}}
  \textbf{\bibinfo{volume}{89}}, \bibinfo{pages}{938--941}
  (\bibinfo{year}{2014}).

\bibitem{world2014antimicrobial}
\bibinfo{author}{Organization, W.~H.} \emph{et~al.}
\newblock \emph{\bibinfo{title}{Antimicrobial resistance: global report on
  surveillance}} (\bibinfo{publisher}{World Health Organization},
  \bibinfo{year}{2014}).

\bibitem{saha2019increasing}
\bibinfo{author}{Saha, S.} \emph{et~al.}
\newblock \bibinfo{title}{Increasing antibiotic resistance in clostridioides
  difficile: a systematic review and meta-analysis}.
\newblock \emph{\bibinfo{journal}{Anaerobe}} \textbf{\bibinfo{volume}{58}},
  \bibinfo{pages}{35--46} (\bibinfo{year}{2019}).

\bibitem{de2018antimicrobial}
\bibinfo{author}{de~Breij, A.} \emph{et~al.}
\newblock \bibinfo{title}{The antimicrobial peptide saap-148 combats
  drug-resistant bacteria and biofilms}.
\newblock \emph{\bibinfo{journal}{Science translational medicine}}
  \textbf{\bibinfo{volume}{10}} (\bibinfo{year}{2018}).

\bibitem{kang2019antimicrobial}
\bibinfo{author}{Kang, J.}, \bibinfo{author}{Dietz, M.~J.} \&
  \bibinfo{author}{Li, B.}
\newblock \bibinfo{title}{Antimicrobial peptide ll-37 is bactericidal against
  staphylococcus aureus biofilms}.
\newblock \emph{\bibinfo{journal}{PLoS One}} \textbf{\bibinfo{volume}{14}},
  \bibinfo{pages}{e0216676} (\bibinfo{year}{2019}).

\bibitem{makowski2019advances}
\bibinfo{author}{Makowski, M.}, \bibinfo{author}{Silva, {\'I}.~C.},
  \bibinfo{author}{Pais~do Amaral, C.}, \bibinfo{author}{Gon{\c{c}}alves, S.}
  \& \bibinfo{author}{Santos, N.~C.}
\newblock \bibinfo{title}{Advances in lipid and metal nanoparticles for
  antimicrobial peptide delivery}.
\newblock \emph{\bibinfo{journal}{Pharmaceutics}}
  \textbf{\bibinfo{volume}{11}}, \bibinfo{pages}{588} (\bibinfo{year}{2019}).

\bibitem{pirtskhalava2016dbaasp}
\bibinfo{author}{Pirtskhalava, M.} \emph{et~al.}
\newblock \bibinfo{title}{Dbaasp v. 2: an enhanced database of structure and
  antimicrobial/cytotoxic activity of natural and synthetic peptides}.
\newblock \emph{\bibinfo{journal}{Nucleic acids research}}
  \textbf{\bibinfo{volume}{44}}, \bibinfo{pages}{D1104--D1112}
  (\bibinfo{year}{2016}).

\bibitem{zhao2013lamp}
\bibinfo{author}{Zhao, X.}, \bibinfo{author}{Wu, H.}, \bibinfo{author}{Lu, H.},
  \bibinfo{author}{Li, G.} \& \bibinfo{author}{Huang, Q.}
\newblock \bibinfo{title}{Lamp: a database linking antimicrobial peptides}.
\newblock \emph{\bibinfo{journal}{PloS one}} \textbf{\bibinfo{volume}{8}},
  \bibinfo{pages}{e66557} (\bibinfo{year}{2013}).

\bibitem{thomas2010camp}
\bibinfo{author}{Thomas, S.}, \bibinfo{author}{Karnik, S.},
  \bibinfo{author}{Barai, R.~S.}, \bibinfo{author}{Jayaraman, V.~K.} \&
  \bibinfo{author}{Idicula-Thomas, S.}
\newblock \bibinfo{title}{Camp: a useful resource for research on antimicrobial
  peptides}.
\newblock \emph{\bibinfo{journal}{Nucleic acids research}}
  \textbf{\bibinfo{volume}{38}}, \bibinfo{pages}{D774--D780}
  (\bibinfo{year}{2010}).

\bibitem{wang2016apd3}
\bibinfo{author}{Wang, G.}, \bibinfo{author}{Li, X.} \& \bibinfo{author}{Wang,
  Z.}
\newblock \bibinfo{title}{Apd3: the antimicrobial peptide database as a tool
  for research and education}.
\newblock \emph{\bibinfo{journal}{Nucleic acids research}}
  \textbf{\bibinfo{volume}{44}}, \bibinfo{pages}{D1087--D1093}
  (\bibinfo{year}{2016}).

\bibitem{seshadri2012dampd}
\bibinfo{author}{Seshadri~Sundararajan, V.} \emph{et~al.}
\newblock \bibinfo{title}{Dampd: a manually curated antimicrobial peptide
  database}.
\newblock \emph{\bibinfo{journal}{Nucleic acids research}}
  \textbf{\bibinfo{volume}{40}}, \bibinfo{pages}{D1108--D1112}
  (\bibinfo{year}{2012}).

\bibitem{das2018pepcvae}
\bibinfo{author}{Das, P.} \emph{et~al.}
\newblock \bibinfo{title}{Pepcvae: Semi-supervised targeted design of
  antimicrobial peptide sequences}.
\newblock \emph{\bibinfo{journal}{arXiv preprint arXiv:1810.07743}}
  (\bibinfo{year}{2018}).

\bibitem{lata2010antibp2}
\bibinfo{author}{Lata, S.}, \bibinfo{author}{Mishra, N.~K.} \&
  \bibinfo{author}{Raghava, G.~P.}
\newblock \bibinfo{title}{Antibp2: improved version of antibacterial peptide
  prediction}.
\newblock \emph{\bibinfo{journal}{BMC bioinformatics}}
  \textbf{\bibinfo{volume}{11}}, \bibinfo{pages}{1--7} (\bibinfo{year}{2010}).

\bibitem{randou2013binary}
\bibinfo{author}{Randou, E.~G.}, \bibinfo{author}{Veltri, D.} \&
  \bibinfo{author}{Shehu, A.}
\newblock \bibinfo{title}{Binary response models for recognition of
  antimicrobial peptides}.
\newblock In \emph{\bibinfo{booktitle}{Proceedings of the International
  Conference on Bioinformatics, Computational Biology and Biomedical
  Informatics}}, \bibinfo{pages}{76--85} (\bibinfo{year}{2013}).

\bibitem{xiao2013iamp}
\bibinfo{author}{Xiao, X.}, \bibinfo{author}{Wang, P.}, \bibinfo{author}{Lin,
  W.-Z.}, \bibinfo{author}{Jia, J.-H.} \& \bibinfo{author}{Chou, K.-C.}
\newblock \bibinfo{title}{iamp-2l: a two-level multi-label classifier for
  identifying antimicrobial peptides and their functional types}.
\newblock \emph{\bibinfo{journal}{Analytical biochemistry}}
  \textbf{\bibinfo{volume}{436}}, \bibinfo{pages}{168--177}
  (\bibinfo{year}{2013}).

\bibitem{fjell2009identification}
\bibinfo{author}{Fjell, C.~D.} \emph{et~al.}
\newblock \bibinfo{title}{Identification of novel antibacterial peptides by
  chemoinformatics and machine learning}.
\newblock \emph{\bibinfo{journal}{Journal of medicinal chemistry}}
  \textbf{\bibinfo{volume}{52}}, \bibinfo{pages}{2006--2015}
  (\bibinfo{year}{2009}).

\bibitem{bhadra2018ampep}
\bibinfo{author}{Bhadra, P.}, \bibinfo{author}{Yan, J.}, \bibinfo{author}{Li,
  J.}, \bibinfo{author}{Fong, S.} \& \bibinfo{author}{Siu, S.~W.}
\newblock \bibinfo{title}{Ampep: Sequence-based prediction of antimicrobial
  peptides using distribution patterns of amino acid properties and random
  forest}.
\newblock \emph{\bibinfo{journal}{Scientific reports}}
  \textbf{\bibinfo{volume}{8}}, \bibinfo{pages}{1--10} (\bibinfo{year}{2018}).

\bibitem{witten2019deep}
\bibinfo{author}{Witten, J.} \& \bibinfo{author}{Witten, Z.}
\newblock \bibinfo{title}{Deep learning regression model for antimicrobial
  peptide design}.
\newblock \emph{\bibinfo{journal}{BioRxiv}} \bibinfo{pages}{692681}
  (\bibinfo{year}{2019}).

\bibitem{das2021accelerated}
\bibinfo{author}{Das, P.} \emph{et~al.}
\newblock \bibinfo{title}{Accelerated antimicrobial discovery via deep
  generative models and molecular dynamics simulations}.
\newblock \emph{\bibinfo{journal}{Nature Biomedical Engineering}}
  \textbf{\bibinfo{volume}{5}}, \bibinfo{pages}{613--623}
  (\bibinfo{year}{2021}).

\bibitem{veltri2018deep}
\bibinfo{author}{Veltri, D.}, \bibinfo{author}{Kamath, U.} \&
  \bibinfo{author}{Shehu, A.}
\newblock \bibinfo{title}{Deep learning improves antimicrobial peptide
  recognition}.
\newblock \emph{\bibinfo{journal}{Bioinformatics}}
  \textbf{\bibinfo{volume}{34}}, \bibinfo{pages}{2740--2747}
  (\bibinfo{year}{2018}).

\bibitem{zou2019mldeepre}
\bibinfo{author}{Zou, Z.}, \bibinfo{author}{Tian, S.}, \bibinfo{author}{Gao,
  X.} \& \bibinfo{author}{Li, Y.}
\newblock \bibinfo{title}{mldeepre: Multi-functional enzyme function prediction
  with hierarchical multi-label deep learning}.
\newblock \emph{\bibinfo{journal}{Frontiers in Genetics}}
  \textbf{\bibinfo{volume}{9}}, \bibinfo{pages}{714} (\bibinfo{year}{2019}).

\bibitem{li2021hmd}
\bibinfo{author}{Li, Y.} \emph{et~al.}
\newblock \bibinfo{title}{Hmd-arg: hierarchical multi-task deep learning for
  annotating antibiotic resistance genes}.
\newblock \emph{\bibinfo{journal}{Microbiome}} \textbf{\bibinfo{volume}{9}},
  \bibinfo{pages}{1--12} (\bibinfo{year}{2021}).

\bibitem{rives2019biological}
\bibinfo{author}{Rives, A.} \emph{et~al.}
\newblock \bibinfo{title}{Biological structure and function emerge from scaling
  unsupervised learning to 250 million protein sequences}.
\newblock \emph{\bibinfo{journal}{bioRxiv}}  (\bibinfo{year}{2019}).
\newblock \urlprefix\url{https://www.biorxiv.org/content/10.1101/622803v4}.

\bibitem{zhou2017deep}
\bibinfo{author}{Zhou, Z.-H.} \& \bibinfo{author}{Feng, J.}
\newblock \bibinfo{title}{Deep forest}.
\newblock \emph{\bibinfo{journal}{arXiv preprint arXiv:1702.08835}}
  (\bibinfo{year}{2017}).

\bibitem{suzek2007uniref}
\bibinfo{author}{Suzek, B.~E.}, \bibinfo{author}{Huang, H.},
  \bibinfo{author}{McGarvey, P.}, \bibinfo{author}{Mazumder, R.} \&
  \bibinfo{author}{Wu, C.~H.}
\newblock \bibinfo{title}{Uniref: comprehensive and non-redundant uniprot
  reference clusters}.
\newblock \emph{\bibinfo{journal}{Bioinformatics}}
  \textbf{\bibinfo{volume}{23}}, \bibinfo{pages}{1282--1288}
  (\bibinfo{year}{2007}).

\bibitem{vaswani2017attention}
\bibinfo{author}{Vaswani, A.} \emph{et~al.}
\newblock \bibinfo{title}{Attention is all you need}.
\newblock In \emph{\bibinfo{booktitle}{Advances in neural information
  processing systems}}, \bibinfo{pages}{5998--6008} (\bibinfo{year}{2017}).

\bibitem{yang2019multi}
\bibinfo{author}{Yang, L.}, \bibinfo{author}{Wu, X.-Z.},
  \bibinfo{author}{Jiang, Y.} \& \bibinfo{author}{Zhou, Z.-H.}
\newblock \bibinfo{title}{Multi-label learning with deep forest}.
\newblock \emph{\bibinfo{journal}{arXiv preprint arXiv:1911.06557}}
  (\bibinfo{year}{2019}).

\bibitem{wu2017unified}
\bibinfo{author}{Wu, X.-Z.} \& \bibinfo{author}{Zhou, Z.-H.}
\newblock \bibinfo{title}{A unified view of multi-label performance measures}.
\newblock In \emph{\bibinfo{booktitle}{International Conference on Machine
  Learning}}, \bibinfo{pages}{3780--3788} (\bibinfo{organization}{PMLR},
  \bibinfo{year}{2017}).

\bibitem{finn2017model}
\bibinfo{author}{Finn, C.}, \bibinfo{author}{Abbeel, P.} \&
  \bibinfo{author}{Levine, S.}
\newblock \bibinfo{title}{Model-agnostic meta-learning for fast adaptation of
  deep networks}.
\newblock In \emph{\bibinfo{booktitle}{International Conference on Machine
  Learning}}, \bibinfo{pages}{1126--1135} (\bibinfo{organization}{PMLR},
  \bibinfo{year}{2017}).

\bibitem{finn2018probabilistic}
\bibinfo{author}{Finn, C.}, \bibinfo{author}{Xu, K.} \&
  \bibinfo{author}{Levine, S.}
\newblock \bibinfo{title}{Probabilistic model-agnostic meta-learning}.
\newblock \emph{\bibinfo{journal}{arXiv preprint arXiv:1806.02817}}
  (\bibinfo{year}{2018}).

\bibitem{grant2018recasting}
\bibinfo{author}{Grant, E.}, \bibinfo{author}{Finn, C.},
  \bibinfo{author}{Levine, S.}, \bibinfo{author}{Darrell, T.} \&
  \bibinfo{author}{Griffiths, T.}
\newblock \bibinfo{title}{Recasting gradient-based meta-learning as
  hierarchical bayes}.
\newblock \emph{\bibinfo{journal}{arXiv preprint arXiv:1801.08930}}
  (\bibinfo{year}{2018}).

\bibitem{gull2019amap}
\bibinfo{author}{Gull, S.}, \bibinfo{author}{Shamim, N.} \&
  \bibinfo{author}{Minhas, F.}
\newblock \bibinfo{title}{Amap: Hierarchical multi-label prediction of
  biologically active and antimicrobial peptides}.
\newblock \emph{\bibinfo{journal}{Computers in biology and medicine}}
  \textbf{\bibinfo{volume}{107}}, \bibinfo{pages}{172--181}
  (\bibinfo{year}{2019}).

\bibitem{chen2016xgboost}
\bibinfo{author}{Chen, T.} \& \bibinfo{author}{Guestrin, C.}
\newblock \bibinfo{title}{Xgboost: A scalable tree boosting system}.
\newblock In \emph{\bibinfo{booktitle}{Proceedings of the 22nd acm sigkdd
  international conference on knowledge discovery and data mining}},
  \bibinfo{pages}{785--794} (\bibinfo{year}{2016}).

\bibitem{ye2020lamp2}
\bibinfo{author}{Ye, G.} \emph{et~al.}
\newblock \bibinfo{title}{Lamp2: a major update of the database linking
  antimicrobial peptides}.
\newblock \emph{\bibinfo{journal}{Database}} \textbf{\bibinfo{volume}{2020}}
  (\bibinfo{year}{2020}).

\bibitem{shi2021dramp}
\bibinfo{author}{Shi, G.} \emph{et~al.}
\newblock \bibinfo{title}{Dramp 3.0: an enhanced comprehensive data repository
  of antimicrobial peptides}.
\newblock \emph{\bibinfo{journal}{Nucleic Acids Research}}
  (\bibinfo{year}{2021}).

\bibitem{madden2013blast}
\bibinfo{author}{Madden, T.}
\newblock \bibinfo{title}{The blast sequence analysis tool}.
\newblock \emph{\bibinfo{journal}{The NCBI handbook}}
  \textbf{\bibinfo{volume}{2}}, \bibinfo{pages}{425--436}
  (\bibinfo{year}{2013}).

\bibitem{uniprot2019uniprot}
\bibinfo{author}{Consortium, U.}
\newblock \bibinfo{title}{Uniprot: a worldwide hub of protein knowledge}.
\newblock \emph{\bibinfo{journal}{Nucleic acids research}}
  \textbf{\bibinfo{volume}{47}}, \bibinfo{pages}{D506--D515}
  (\bibinfo{year}{2019}).

\bibitem{fu2012cd}
\bibinfo{author}{Fu, L.}, \bibinfo{author}{Niu, B.}, \bibinfo{author}{Zhu, Z.},
  \bibinfo{author}{Wu, S.} \& \bibinfo{author}{Li, W.}
\newblock \bibinfo{title}{Cd-hit: accelerated for clustering the
  next-generation sequencing data}.
\newblock \emph{\bibinfo{journal}{Bioinformatics}}
  \textbf{\bibinfo{volume}{28}}, \bibinfo{pages}{3150--3152}
  (\bibinfo{year}{2012}).

\bibitem{breiman2001random}
\bibinfo{author}{Breiman, L.}
\newblock \bibinfo{title}{Random forests}.
\newblock \emph{\bibinfo{journal}{Machine learning}}
  \textbf{\bibinfo{volume}{45}}, \bibinfo{pages}{5--32} (\bibinfo{year}{2001}).

\bibitem{van2008visualizing}
\bibinfo{author}{Van~der Maaten, L.} \& \bibinfo{author}{Hinton, G.}
\newblock \bibinfo{title}{Visualizing data using t-sne.}
\newblock \emph{\bibinfo{journal}{Journal of machine learning research}}
  \textbf{\bibinfo{volume}{9}} (\bibinfo{year}{2008}).

\bibitem{ribeiro2016should}
\bibinfo{author}{Ribeiro, M.~T.}, \bibinfo{author}{Singh, S.} \&
  \bibinfo{author}{Guestrin, C.}
\newblock \bibinfo{title}{" why should i trust you?" explaining the predictions
  of any classifier}.
\newblock In \emph{\bibinfo{booktitle}{Proceedings of the 22nd ACM SIGKDD
  international conference on knowledge discovery and data mining}},
  \bibinfo{pages}{1135--1144} (\bibinfo{year}{2016}).

\bibitem{burdukiewicz2020proteomic}
\bibinfo{author}{Burdukiewicz, M.} \emph{et~al.}
\newblock \bibinfo{title}{Proteomic screening for prediction and design of
  antimicrobial peptides with ampgram}.
\newblock \emph{\bibinfo{journal}{International journal of molecular sciences}}
  \textbf{\bibinfo{volume}{21}}, \bibinfo{pages}{4310} (\bibinfo{year}{2020}).

\bibitem{torrent2012ampa}
\bibinfo{author}{Torrent, M.} \emph{et~al.}
\newblock \bibinfo{title}{Ampa: an automated web server for prediction of
  protein antimicrobial regions}.
\newblock \emph{\bibinfo{journal}{Bioinformatics}}
  \textbf{\bibinfo{volume}{28}}, \bibinfo{pages}{130--131}
  (\bibinfo{year}{2012}).

\bibitem{waghu2016campr3}
\bibinfo{author}{Waghu, F.~H.}, \bibinfo{author}{Barai, R.~S.},
  \bibinfo{author}{Gurung, P.} \& \bibinfo{author}{Idicula-Thomas, S.}
\newblock \bibinfo{title}{Campr3: a database on sequences, structures and
  signatures of antimicrobial peptides}.
\newblock \emph{\bibinfo{journal}{Nucleic acids research}}
  \textbf{\bibinfo{volume}{44}}, \bibinfo{pages}{D1094--D1097}
  (\bibinfo{year}{2016}).

\bibitem{lam2019deep}
\bibinfo{author}{Lam, J.~H.} \emph{et~al.}
\newblock \bibinfo{title}{A deep learning framework to predict binding
  preference of rna constituents on protein surface}.
\newblock \emph{\bibinfo{journal}{Nature communications}}
  \textbf{\bibinfo{volume}{10}}, \bibinfo{pages}{1--13} (\bibinfo{year}{2019}).

\bibitem{wei2021protein}
\bibinfo{author}{Wei, J.}, \bibinfo{author}{Chen, S.}, \bibinfo{author}{Zong,
  L.}, \bibinfo{author}{Gao, X.} \& \bibinfo{author}{Li, Y.}
\newblock \bibinfo{title}{Protein-rna interaction prediction with deep
  learning: Structure matters}.
\newblock \emph{\bibinfo{journal}{arXiv preprint arXiv:2107.12243}}
  (\bibinfo{year}{2021}).

\bibitem{delano2002pymol}
\bibinfo{author}{DeLano, W.~L.} \emph{et~al.}
\newblock \bibinfo{title}{Pymol: An open-source molecular graphics tool}.
\newblock \emph{\bibinfo{journal}{CCP4 Newsletter on protein crystallography}}
  \textbf{\bibinfo{volume}{40}}, \bibinfo{pages}{82--92}
  (\bibinfo{year}{2002}).

\end{thebibliography}

\newpage
  \section*{Appendix}
\appendix
\renewcommand{\thetable}{\Alph{section}.1}

\section{Structure investigation of the novel AMPs}
We draw the 3D structure graphs of YI12 and FK13, and find that ‘YLR’ subsequence in YI12 and ‘WLK’ subsequence in FK13 (stick model in Figure \ref{F7}) can be aligned, and the RMSD (Root-mean-square deviation of atomic positions) score is 0.111. We suspect that these three loci make two sequences resistant to gram-positive and gram-negative. However, many existing loci prediction tools failed to predict YI12 as an AMP, and we are unable to validate the above conjecture with the existing computational tools. In the future, we will try to develop more accurate generative models and loci prediction models based on our proposed method, discovering new AMPs and identifying their functional loci.  
\begin{table*}[!h]
\centering
\caption{Sequences of the 20 peptides.}
\label{table 6}
\scalebox{0.8}{
\begin{tabular}{ll}
\toprule
Description & Sequence\\
\midrule
\textbf{YI12} & \textbf{YLRLIRYMAKMI}\\
\textbf{FK13} & \textbf{FPLTWLKWWKWKK}\\
- & HILRMRIRQMMT\\
- & ILLHAILGVRKKL\\
- & YRAAMLRRQYMMT\\
- & HIRLMRIRQMMT\\
- & HIRAMRIRAQMMT\\
- & KTLAQLSAGVKRWH\\
- & HILRMRIRQGMMT\\
- & HRAIMLRIRQMMT\\
- & EYLIEVRESAKMTQ\\
- & GLITMLKVGLAKVQ\\
- & YQLLRIMRINIA\\
- & VRWIEYWREKWRT\\
- & LIQVAPLGRLLKRR\\
- & YQLRLIMKYAI\\
- & HRALMRIRQCMT\\
- & GWLPTEKWRKLC\\
- & YQLRLMRIMSRI\\
- & LRPAFKVSK\\
\bottomrule
\end{tabular}}
\end{table*}

\renewcommand{\thefigure}{\Alph{section}.1}
\begin{figure}[!htbp] 
\centering 
\includegraphics[width=0.8\textwidth]{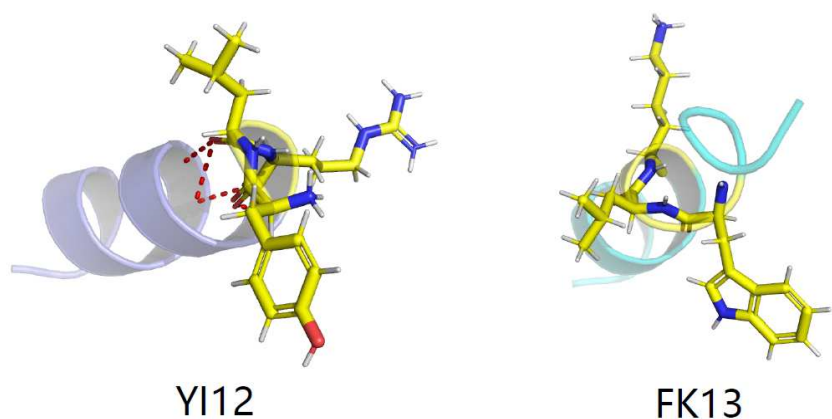} 
\caption{3D structure graphs of YI12 and FK13 generated by PyMOL \protect\cite{delano2002pymol}} 
 
\label{F7} 
\end{figure}

\renewcommand{\thefigure}{\Alph{section}.1}
\section{Feature analysis}
\begin{figure}[h] 
\centering 
\includegraphics[width=1.0\textwidth]{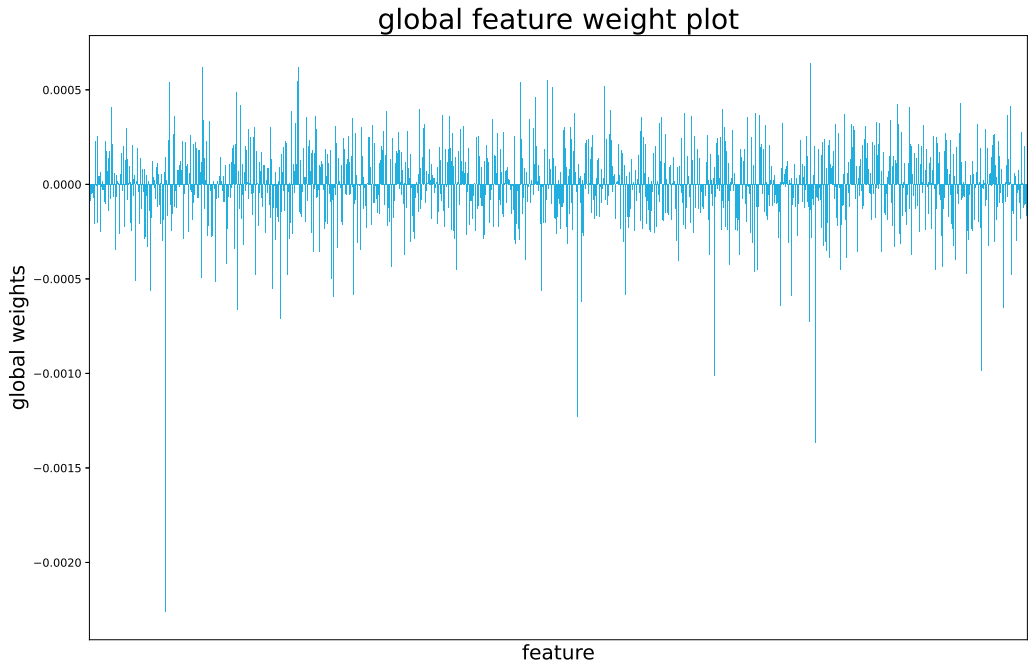} 
\caption{Global weights of 1280 features. We get 1280 features' global weights in the classification task by applying LIME. Features with weights greater than 0 have positive effects on prediction, while features with weights less than 0 have negative effects.} 
 
\label{F4} 
\end{figure}

We adopt Local Interpretable Model-agnostic Explanations (LIME) \cite{ribeiro2016should}. And LIME is an explanation technique that explains the prediction of our classifier in an interpretable and faithful manner by learning an interpretable model locally around the prediction. We obtain 1280 features' effect on the prediction (global weights) by using the LIME framework (Figure \ref{F4}). And those global weights of features are predicted based on the average value of local prediction. In fact, the main function of LIME framework is to find features that have the most positive impact on the model. And these selected features could help the model better fit the given data. We select 48 features (Figure \ref{F5}) with the highest weight, and test whether the model trained with these features performs well or not.

\renewcommand{\thefigure}{\Alph{section}.2}

\begin{figure}[ht] 
\centering 
\includegraphics[width=1.0\textwidth]{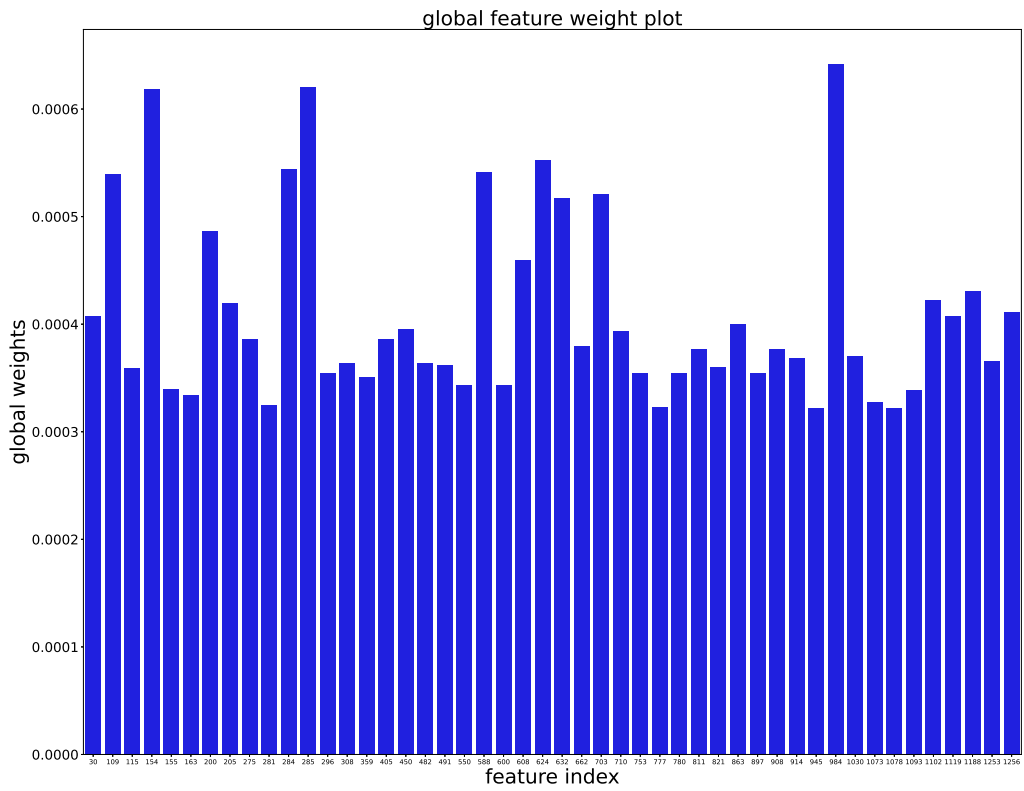} 
\caption{We select 48 features with the highest weight among 1280 features. The X-axis shows indexes of 48 features. These features are used to train our function prediction model, in order to achieve faster training speed and less computational resource consumption.} 
\label{F5} 
\end{figure}

\end{document}